\documentclass[10pt,twocolumn,letterpaper]{article}

\usepackage{cvpr}              
\usepackage{multirow}
\usepackage[dvipsnames]{xcolor}
\usepackage{caption}
\usepackage{colortbl}
\usepackage{xcolor} 
\newcommand{\cc}[1]{\cellcolor{gray!20}#1}
\usepackage{tabularx}
\usepackage{booktabs}    
\usepackage{multirow}    
\usepackage{graphicx}    

\definecolor{cvprblue}{rgb}{0.21,0.49,0.74}
\usepackage[pagebackref,breaklinks,colorlinks,allcolors=cvprblue]{hyperref}

\def\paperID{13186} 
\def\confName{CVPR}
\def\confYear{2026}

\title{\vspace{-5pt}\raisebox{-0.5em}{\includegraphics[height=2em]{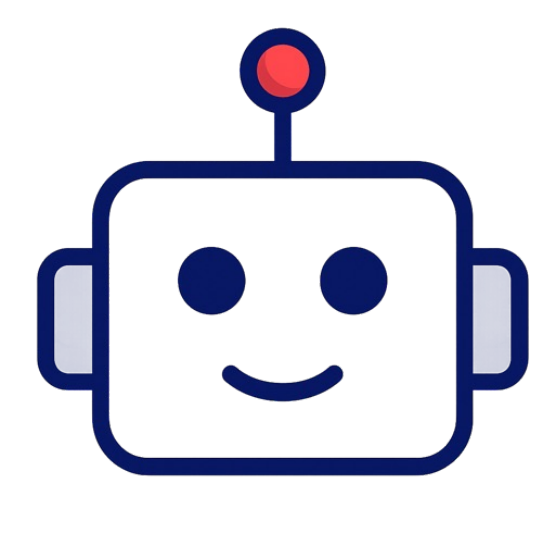}}\ \  Robo-SGG: Exploiting Layout-Oriented Normalization and Restitution Can Improve Robust Scene Graph Generation}

\author{Changsheng Lv \qquad Zijian Fu \qquad Mengshi Qi\thanks{Corresponding author: qms@bupt.edu.cn.} \\
 State Key Laboratory of Networking and Switching Technology, \\ Beijing University of Posts and Telecommunications, China\\
}

\begin{document}
\maketitle
\begin{abstract}
In this paper, we propose Robo-SGG, a plug-and-play module for robust scene graph generation (SGG). Unlike standard SGG, the robust scene graph generation aims to perform inference on a diverse range of corrupted images, with the core challenge being the domain shift between the clean and corrupted images. Existing SGG methods suffer from degraded performance due to shifted visual features (\textit{e.g.}, corruption interference or occlusions). To obtain robust visual features, we leverage layout information—representing the global structure of an image—which is robust to domain-shift, to enhance the robustness of SGG methods under corruption. Specifically, we employ Instance Normalization~(IN) to alleviate the domain-specific variations and recover the robust structural features~(\textit{i.e.}, the positional and semantic relationships among objects) by the proposed Layout-Oriented Restitution. Furthermore, under corrupted images, we introduce a Layout-Embedded Encoder (LEE) that adaptively fuses layout and visual features via a gating mechanism, enhancing the robustness of positional and semantic representations for objects and predicates. Note that our proposed Robo-SGG module is designed as a plug-and-play component, which can be easily integrated into any baseline SGG model. Extensive experiments demonstrate that by integrating the state-of-the-art method into our proposed Robo-SGG, we achieve relative improvements of 6.3\%, 11.1\%, and 8.0\% in mR@50 for PredCls, SGCls, and SGDet tasks on the VG-C benchmark, respectively, and achieve new state-of-the-art performance in the corruption scene graph generation benchmark~(VG-C and GQA-C). We will release our source code and model.
\end{abstract}    
\begin{center}
\vspace{-3mm}
\textit{``The power of geometry lies in transforming chaotic intuition into clear necessity.''}\\
\hfill --- Bertrand Russell
\vspace{-3mm}
\end{center}
\section{Introduction}
\label{sec:intro}

Scene Graph Generation~(SGG)~\cite{xu2017scene,yu2017visual,li2017scene, yang2018graph} has been introduced to create a visually-grounded graph, wherein the nodes represent detected object instances and the edges encapsulate their pairwise relationships. The SGG model has been widely used in many vision tasks, such as autonomous driving~\cite{wang2024rs2g, zhang2024graphad,ye2025safedriverag,lv2025t2sg,zhu2023unsupervised,qi2020stc} and robotic navigation~\cite{werby2024hierarchical,singh2023scene,deng2025global,qi2021semantics}.

However, most existing studies in this field assume access to ``clean'' images, but in real-world scenarios, one encounters various types of impaired data (natural~\cite{hendrycks2019benchmarking}, adversarial~\cite{ito2021optimal}, etc.), with the most prevalent being corruptions that commonly arise from natural phenomena such as noise, blur, various weather conditions, and digital distortions~\cite{hendrycks2019benchmarking}, which significantly limits their practical applicability. This highlights the necessity for SGG models to be robust and capable of handling such corruption effectively. In our paper, we focus on the SGG task under image corruptions arising from natural phenomena.

\begin{figure}[t]
    \centering
    \includegraphics[width=1\linewidth]{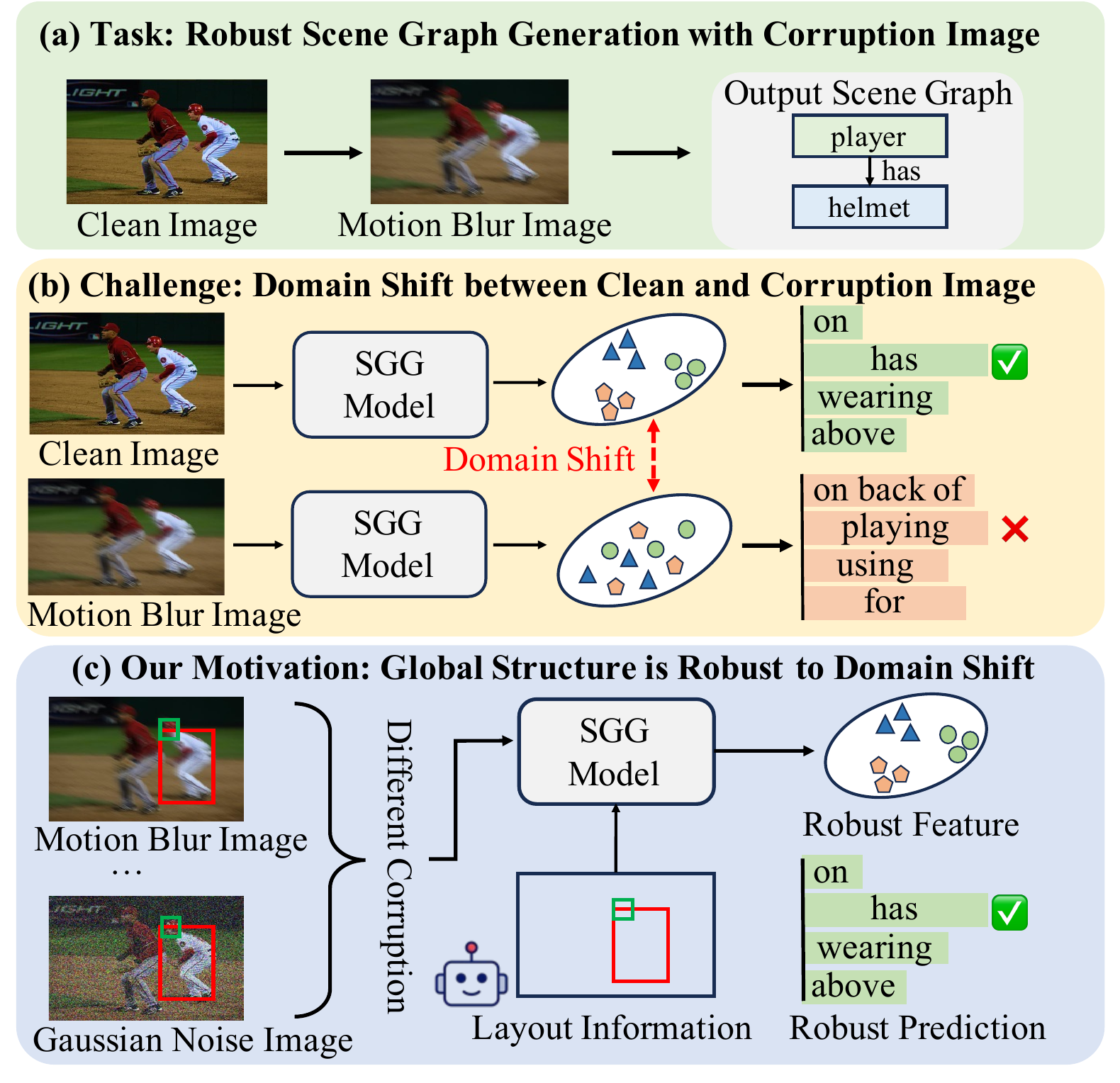}
    \vspace{-7mm}
    \caption{(a) Illustration of the robust SGG task. (b) Feature domain shift between clean and corrupted image features degrades model performance. (c) Robo-SGG leverages layout information to improve the robustness of structural features and object/predicate representations.}
    \label{fig: an overview of our model}
    \vspace{-6mm}
\end{figure} 

Existing scene graph generation methods~\cite{dong2022stacked,li2024leveraging,zhang2024hiker} leverage visual, textual, and external knowledge graph features to enhance model generalization via multimodal interaction. However, under the interference of corruption, visual features undergo feature domain shifts~\cite{huang2023normalization}, which erroneously guide the interaction among multimodal information, thereby significantly degrading the performance of SGG models.  To improve domain generalization under such shifts, current robust models adopt strategies like data augmentation~\cite{hendrycks2019augmix, zhang2022memo}, adversarial training~\cite{herrmann2022pyramid,kireev2022effectiveness,rusak2020simple}, normalization~\cite{zhou2019omni,wang2020tent}, and additional denoising~\cite{garber2024image} to perform well on both seen and unseen domains with various corruptions. However, these methods introduce considerable computational overhead and are mainly object-centric, failing to enhance the robustness of structural features ({\it i.e.}, positional and semantic relationships among objects). In contrast, layout information, which captures the global spatial arrangement of objects, is inherently more robust to domain shifts than low-level appearance cues such as texture or color~\cite{binmakhashen2019document}. Consequently, leveraging robust layout representations is particularly well-suited for SGG, which requires comprehensive semantic relationship recognition.

To address this limitation, we propose leveraging the image layout, which captures global structure and is less vulnerable to corruption than low-level features like texture and color, via a newly designed Layout-Oriented Normalization and Restitution Module (NRM) for more robust structural features (see Figure~\ref{fig: an overview of our model}~(c)). The key challenge for NRM is to alleviate domain-specific features while retaining robust structural features across different corruptions. We first apply Instance Normalization (IN) to alleviate corruption-specific mean and variance, thereby alleviating domain-specific variations. Subsequently, we propose a layout-aware attention mechanism to capture the image's structure and recover domain-generalizable structural features based on the attention weights derived from the residual feature, which represents the discrepancy between the original and normalized information.

Furthermore, we propose the Layout-Embedded Encoder (LEE) to obtain robust object and predicate features from individual object bounding boxes and their interactions. Although the existing method SHA~\cite{dong2022stacked} explores the fusion of object and predicate spatial and visual information via concatenation, it relies on reliable detection results, which are often inaccurate and noisy under corruptions. In contrast, our proposed LEE learns to downweight unreliable spatial cues and adaptively balances visual and spatial information for robust object and predicate representation under corruption. Specifically, for objects, we use their visual features to obtain gating coefficients that flexibly fuse the embeddings of bounding box coordinates and visual features. For predicates (\textit{i.e.}, object interactions), we similarly use the visual features extracted from the union region of the object pair as conditions to generate gating coefficients for fusing the embeddings of the object pair's bounding box coordinates and visual features.

Our main contributions can be summarized as follows:
\par\textbf{(1)} We propose a novel approach, Robo-SGG, for the robust scene graph generation by designing a Layout-Oriented Normalization and Restitution Module~(NRM) to attain a generalized structural feature.

\par\textbf{(2)} We design a new Layout-Embedded Encoder (LEE), which adaptively fuses the embedding of object bounding box coordinates and visual features via gating to obtain robust object and predicate representations under corruption.

\par\textbf{(3)} Experimental results demonstrate our proposed Robo-SGG is a plug-and-play method that can be incorporated into any current SGG baselines and show it can achieve relative improvements of 6.3\%, 11.1\%, and 8.0\% in mR@50 for PredCls, SGCls, and SGDet tasks in the VG-C benchmarks, respectively. 
\section{Related Work}
\label{sec: Related Work}
\begin{figure*}[t!]
    \centering
    \includegraphics[width=1.0\linewidth]{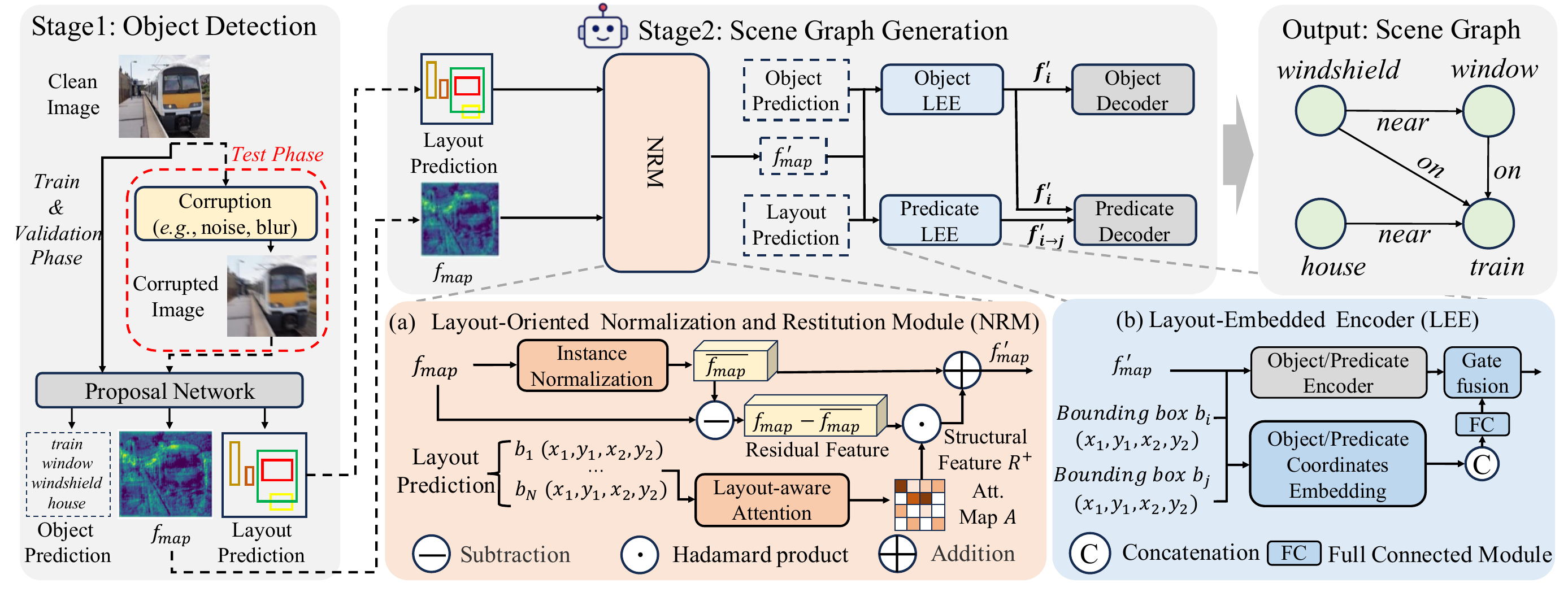}
    \vspace{-6mm}
   \caption{{\bf Overall framework of our proposed Robo-SGG.} Stage 1, Stage 2, and Output denote the standard SGG pipeline, with our NRM and LEE modules integrated into Stage 2. In Stage 1, only clean images are used during training, while corrupted images are employed during validation and testing. Illustrated with a two-stage SGG model: (a)~NRM uses Instance Normalization and layout-aware attention to alleviate domain disturbances and restore robust structural features; (b)~LEE fuses visual features and bounding box coordinates embedding via gated fusion for robust object and predicate representations.}
    \label{fig: Main Model}
    \vspace{-5mm}
\end{figure*}

\noindent\textbf{Scene Graph Generation.}
Scene graphs, originally conceptualized for image retrieval, deconstruct an image into its fundamental elements: objects, their attributes, and the relationships among them~\cite{johnson2015image}. The availability of large-scale datasets like Visual Genome (VG)~\cite{krishna2017visual} and GQA~\cite{hudson2019gqa} has driven significant progress in scene graph generation (SGG)~\cite{zheng2023prototype, liu2021fully, chen2024expanding,qi2019attentive}. Recent research addresses challenges such as label bias~\cite{tang2020unbiased, sun2023unbiased,11222969,11185304,11223230}, external knowledge integration~\cite{zareian2020bridging, chen2023more, zareian2020learning}, semantic relationship modeling~\cite{zheng2023prototype, zhang2023learning}, and causal reasoning~\cite{tang2020unbiased, sun2023unbiased}, using both one-stage~\cite{im2024egtr, chen2024expanding} and two-stage~\cite{zellers2018neural, tang2019learning} framework. The emergence of visual language models (VLMs) has further advanced open-vocabulary SGG~\cite{li2024pixels, kim2024llm4sgg}. However, most methods assume reliable object detector outputs, which are often unrealistic in the presence of image corruptions. To address this, we enhance object and predicate representations by adaptively balancing detected visual and spatial information, thereby improving SGG robustness in real-world scenarios. 

\noindent\textbf{Robustness in Scene Understanding.}
While most scene understanding models assume clean images, real-world data often contains corruptions that degrade performance. Some works target specific corruptions~\cite{lin2024improving, garber2024image}, but general robustness remains challenging. Benchmarks like ImageNet-C~\cite{hendrycks2019benchmarking} evaluate robustness across diverse corruption types. Existing approaches mainly focus on domain adaptation~\cite{mirza2022norm, gao2023back, benz2021revisiting} and feature generalization~\cite{mintun2021interaction, yin2019fourier, zhang2022memo}, yet often overlook corruption’s impact on inter-object structural features. HiKER~\cite{zhang2024hiker} recently introduced SGG corruption benchmarks and a knowledge-based robustness method, but its reliance on external knowledge graphs limits flexibility. In contrast, Robo-SGG learns robust, layout-aware structural features that mitigate domain shift caused by corruptions, enabling robust, transferable SGG on corrupted images without extra knowledge graphs.
\section{Proposed Method}

\subsection{Problem Definition}
The task of scene graph generation aims to generate a structured representation graph~$\mathcal{G}=( \mathcal{O}, \mathcal{E})$ of the given image~$\mathcal{I}$. Each node~$o_i\in \mathcal{O}$ in a scene graph represents an object with bounding boxe~$b_i=(x_1, y_1, x_2, y_2)$ and the corresponding categorie~$c_i$, while each edge~$p_{i \rightarrow j} \in \mathcal{E}$ correspond to the predicate categorie between the object pair $(o_i, o_j)$. The output of the SGG method is a collection of visual triplets $\langle \text{subject}, \text{predicate}, \text{object} \rangle$, which can be formulated as $\{ (o_i, p_{i \rightarrow j}, o_j) \mid o_i, o_j \in \mathcal{O},\ p_{i \rightarrow j} \in \mathcal{E} \}$.

\subsection{Standard SGG Framework}
As shown in Figure~\ref{fig: Main Model}, our framework is based on the standard SGG pipeline.  Notably, our Robo-SGG can be flexibly added to both one-stage~\cite{cong2023reltr} and two-stage~\cite{zellers2018neural} SGG models, which generally contain three parts: {\bf Proposal Network}, {\bf Object/Predicate Encoder}, and {\bf Object/Predicate Decoder}. The integration of our method into one-stage and two-stage models differs slightly, as discussed in the Proposal Network. Unless otherwise specified, we use the two-stage model as the example in the following description.

\noindent {\bf Proposal Network.} We employ Faster R-CNN~\cite{ren2016faster} with ResNet-101~\cite{he2016deep} as the backbone $\phi_f$ to extract multi-level feature maps $f_{map}$ from the input image $\mathcal{I}$. For simplicity, we illustrate our method using a single feature map layer:
\begin{equation}
    f_{map} = \phi_{f}(\mathcal{I}),
\label{Eq.1}
\end{equation}
where $f_{map} \in \mathbb{R}^{C \times H \times W}$, and $C$, $H$, $W$ denote the number of channels, height, and width, respectively. In two-stage SGG models, the proposal network outputs $\mathcal{O} = \{o_i\}_{i=1}^N$, where $N$ is the number of detected objects and $o_i = (v_i, b_i, c_i)$ represents the visual feature, bounding box, and category of each object. For one-stage SGG models, our framework utilizes the object proposals and features predicted by the Transformer decoder~\cite{cong2023reltr}.

\noindent {\bf Object/Predicate Encoder} is used to embed object and predicate features for subsequent modeling and prediction, as follows:
\begin{align}
    f_i &= \mathrm{Enc}^{obj}\left([v_i,\, \mathrm{Emb}^{lang}(c_i),\, \mathrm{Emb}^{bbox}(b_i)]\right) \label{Eq.2} \\
    f_{i \rightarrow j} &= \mathrm{Enc}^{pred}\left([f_i,\, v_{i \rightarrow j},\, f_j]\right) \label{Eq.3}
\end{align}
where $v_i$ denotes the visual feature of the $i$-th object, and $v_{i \rightarrow j}$ denotes the visual feature of the union region of objects $i$ and $j$ in the feature map $f_{map}$. $\mathrm{Enc}^{obj}$ and $\mathrm{Enc}^{pred}$ denote the object and predicate encoders, which can adopt various encoding modules (\textit{e.g.}, BiLSTM~\cite{zellers2018neural}). The operation $[\cdot, \cdot]$ denotes feature concatenation along the feature dimension. $\mathrm{Emb}^{lang}(\cdot)$ embeds the object category $c_i$ into a textual feature vector using a pre-trained language model (\textit{e.g.}, GloVe~\cite{pennington2014glove}), while $\mathrm{Emb}^{bbox}(b_i)$ embeds the bounding box $b_i$ into a vector via a fully connected embedding module. Finally, $f_i$ and $f_{i \rightarrow j}$ represent the encoded feature of the $i$-th object and the encoded predicate feature between objects $i$ and $j$, respectively.

\noindent \textbf{Object/Predicate Decoder} predicts the final object label $c_i^{\prime}$ and predicate label $p_{i \rightarrow j}$ from the object feature $f_i$ and predicate feature $f_{i \rightarrow j}$, as formulated below:
\begin{align}
    c_i^{\prime}      &= \arg\max(\mathrm{Softmax}(\mathrm{Dec}^{obj}(f_i))), \\
    p_{i \rightarrow j} &= \arg\max(\mathrm{Softmax}(\mathrm{Dec}^{pred}(f_{i \rightarrow j}))),
\end{align}
where $\mathrm{Dec}^{obj}$ and $\mathrm{Dec}^{pred}$ denote the object and predicate decoders, respectively, which are typically implemented as a single fully connected module.

\subsection{Layout-Oriented Normalization and Restitution Module~(NRM)}
As depicted in Figure~\ref{fig: Main Model}(a), we propose the Layout-Oriented Normalization and Restitution Module (NRM), which takes as input the layout information (\textit{i.e.}, all bounding boxes of objects in image~$\mathcal{I}$) and the feature map $f_{map}$ from the backbone $\phi_{f}$, and outputs an enhanced feature map $f_{map}^{\prime}$ that preserves structural information and is robust to various corruptions.  {\bf Our key insight} is that various forms of corruption induce a covariate shift within the feature domain~\cite{huang2023normalization}, with each type of corruption representing a distinct domain. Consequently, our motivation is to utilize Instance Normalization to alleviate domain-specific disturbances and to enhance the generalization of the structural features in $f_{map}$ across different domains by leveraging the overall structural information, which is less vulnerable to corruption than low-level information like texture and color. The NRM consists of two parts: Instance Normalization and Layout-aware Attention.

\noindent {\bf Instance Normalization~(IN).} To address the domain shift in the $f_{map}$ between clean and corrupted images, we employ Instance Normalization (IN)~\cite{ulyanov2016instance} to alleviate the domain-specific disturbance from the feature map $f_{map}$. Specifically, for each image, we compute the mean and variance for each channel:
\begin{equation}
    \mu_{i} = \frac{1}{HW} \sum_{m=1}^{H} \sum_{l=1}^{W} a_{i m l},, 
    \sigma_{i}^2 = \frac{1}{HW} \sum_{m=1}^{H} \sum_{l=1}^{W} (a_{i m l} - \mu_{i})^2,
\end{equation}
where $a_{i m l}$ is the value at channel $i$ and spatial location $(m, l)$ in $f_{map} \in \mathbb{R}^{C \times H \times W}$. Notably, unlike Batch Normalization~(BN)~\cite{li2016revisiting}, which computes statistics across the batch and spatial dimensions, IN computes the mean and variance for each channel of each image over the spatial dimensions only. 
This approach can effectively alleviate the covariance shift caused by feature domain shift~\cite{ulyanov2016instance}. 
The instance-normalized feature map $\overline{f_{map}}$ is then computed as:
\begin{equation}
    b_{i m l} = \frac{a_{i m l} - \mu_{i}}{\sqrt{\sigma_{i}^2 + \epsilon}},
\end{equation}
where $a_{i m l}$ and $b_{i m l}$ are the values before and after IN, respectively. The residual feature $R$ is computed as:
\begin{equation}
    R = f_{map} - \overline{f_{map}}.
\end{equation}

\noindent {\bf Layout-aware Attention.} While IN alleviate domain-specific disturbances in $f_{map}$, it may also remove structural features important for SGG. Therefore, we introduce Layout-aware Attention, which utilizes the coordinates of all object bounding boxes to model the global layout and structural patterns within the image. Specifically, let $e_j = (x_j, y_j)$ denote the centroid of the $j$-th object ($j=1,2,\ldots,N$). Both the spatial locations $(m, l)$ in the feature map and the object centroids $(x_j, y_j)$ are normalized to the range $[0, 1]$ by dividing by the image width and height, respectively. For each normalized location $(m, l)$, the attention weight toward object $j$ is defined as:
\begin{equation}
    A_{(m,l),j} = \frac{\exp\!\left(-\| (m, l) - e_j \|^2\right)}{\sum_{j'=1}^N \exp\!\left(-\| (m, l) - e_{j'} \|^2\right)},
    \label{Eq.9}
\end{equation}
where $A \in \mathbb{R}^{(H \times W) \times N}$. We then derive a layout mask $M \in \mathbb{R}^{H\times W}$ by taking the maximum attention weight across objects at each location, and the structural feature $R^+ \in \mathbb{R}^{C \times H \times W} $ is denoted:
\begin{equation}
    M_{m,l} = \max_{j} A_{(m,l),j},  R^+ = R \odot M,
\end{equation}
where $\odot$ denotes broadcasting multiplication across channels.
The final output feature map is obtained by combining the instance-normalized feature $\overline{f_{map}}$ and the structural feature $R^+$:
\begin{equation}
    f_{map}^{\prime} = \overline{f_{map}} + R^+.
\end{equation}
This output effectively suppresses domain-specific disturbance while retaining essential structural information.

\subsection{Layout-Embedded Encoder~(LEE)}
As depicted in Figure~\ref{fig: Main Model}(b), we propose the Object/Predicate Layout-Embedded Encoder (LEE) to enhance the original object and predicate encoders described in Eq.~\ref{Eq.2} and Eq.~\ref{Eq.3}. LEE consists of two components: Object LEE and Predicate LEE. It establishes a more robust interaction between visual features and bounding box coordinate embeddings, especially under corrupted conditions.

\noindent {\bf Object/Predicate Coordinate Embedding} has been widely adopted in previous works~\cite{dong2022stacked}, where bounding box coordinates are embedded and concatenated with visual features. Specifically, for an object $o_i$ with bounding box $b_i$ and an object pair $(o_i, o_j)$, the coordinate embeddings are defined as:
\begin{align}
    f_i^C &= \mathrm{Emb}^{obj-bbox}(b_i), \\
    f_{i \rightarrow j}^C &= \mathrm{Emb}^{pred-bbox}([b_i, b_j, e_i - e_j, \|b_i - b_j\|_2]),
\end{align}
where $\mathrm{Emb}^{obj-bbox}(\cdot)$ and $\mathrm{Emb}^{pred-bbox}(\cdot)$ are two-layer fully connected embedding modules, $e_i$ is the center of $b_i$, and $[\cdot, \cdot]$ denotes feature concatenation.

While such coordinate embeddings are effective when the detector provides accurate results, they may introduce noise when the detector outputs unreliable bounding boxes under corrupted images. To address this, we introduce a \textbf{Gate Fusion} mechanism that adaptively balances the contributions of visual feature and coordinate embeddings, thereby reducing the impact of inaccurate bounding boxes on SGG performance.

\noindent  {\bf Gate Fusion.} We fuse object/predicate features $f_i, f_{i \rightarrow j} \in \mathbb{R}^d$ with their coordinate embeddings $f_i^C, f_{i \rightarrow j}^C \in \mathbb{R}^d$ via gating. For objects, we compute a gate coefficient $z_i \in \mathbb{R}^d$ as:
\begin{equation}
    z_i = \text{Sigmoid}(W f_i),
    \label{Eq.17}
\end{equation}
where \( W \in \mathbb{R}^{d \times d} \) is a learnable parameter, and $\text{Sigmoid}$ maps the gate coefficient to $[0, 1]$, indicating the proportion of \( f_i^C \) to be retained. The fused feature is then:
\begin{equation}
    f_i^\prime = (1 - z_i) \circ f_i^C + z_i \circ f_i,
    \label{Eq.18}
\end{equation}
with $\circ$ denoting element-wise multiplication. The same process applies to predicates, yielding $f_{i \rightarrow j}^\prime$. No dimensionality reduction or projection is used, and all features are aligned in the same space $d$. 

\begin{table*}[h]
\centering
\small
\setlength{\tabcolsep}{2.5pt}
\renewcommand{\arraystretch}{1}
\scalebox{0.95}{
\begin{tabular}{ll@{\hskip 3pt}ccc@{\hskip 3pt}ccc@{\hskip 3pt}ccc}
\toprule
                         & \multirow{2}{*}{Method} & \multicolumn{3}{c}{PredCls} & \multicolumn{3}{c}{SGCls} & \multicolumn{3}{c}{SGDet} \\
                         &                         & Clean$\uparrow$ & Corruption Avg.$\uparrow$ & Imp.$\uparrow$ & Clean$\uparrow$ & Corruption Avg.$\uparrow$ & Imp.$\uparrow$ & Clean$\uparrow$ & Corruption Avg.$\uparrow$ & Imp.$\uparrow$ \\
\midrule
\multirow{14}{*}{\centering \rotatebox{90}{mR@50}}  
& MOTIFS~\cite{zellers2018neural}         & 14.6 & 12.6 & - & 11.3 & 4.5 & - & 7.2 & 2.8 & - \\
& \cc MOTIFS+Ours                         & \cc 14.6 & \cc 13.2 & \cc \textcolor{black}{\bf +4.8\%} & \cc 11.4 & \cc 4.9 & \cc \textcolor{black}{\bf +8.8\%} & \cc 7.1 & \cc 2.9 & \cc \textcolor{black}{\bf +3.6\%} \\
& VCTree~\cite{tang2019learning}          & 14.9 & 12.8 & - & 12.6 & 5.4 & - & 7.3 & 2.5 & - \\ 
& \cc VCTree+Ours                         & \cc 15.3 & \cc 13.6 & \cc \textcolor{black}{\bf +6.3\%} & \cc 13.3 & \cc 6.0 & \cc \textcolor{black}{\bf +11.1\%} & \cc 7.3 & \cc 2.7 & \cc \textcolor{black}{\bf +8.3\%} \\
& VTransE~\cite{tang2020unbiased}         & 14.7 & 12.1 & - & 12.6 & 4.6 & - & 7.7 & 2.9 & - \\ 
& \cc VTransE+Ours                        & \cc 14.9 & \cc 13.2 & \cc \textcolor{black}{\bf +9.0\%} & \cc 12.4 & \cc 4.9 & \cc \textcolor{black}{\bf +6.5\%} & \cc 7.8 & \cc 3.0 & \cc \textcolor{black}{\bf +3.4\%} \\
& HiKER~\cite{zhang2024hiker}             & 39.3 & 32.6 & - & 20.3 & 3.5 & - & - & - & - \\ 
& \cc HiKER+Ours                          & \cc 40.8 & \cc 33.8 & \cc \textcolor{black}{\bf +3.7\%} & \cc 21.5 & \cc 3.7 & \cc \textcolor{black}{\bf +5.7\%} & \cc - & \cc - & \cc - \\
& DPL~\cite{jeon2024semantic}             & 25.5 & 24.5 & - & 17.7 & 7.2 & - & 12.8 & 4.8 & - \\ 
& \cc DPL+Ours                            & \cc 25.7 & \cc 24.8 & \cc \textcolor{black}{\bf +1.1\%} & \cc 19.4 & \cc 7.8 & \cc \textcolor{black}{\bf +8.3\%} & \cc 13.1 & \cc 5.1 & \cc \textcolor{black}{\bf +6.3\%} \\
\midrule
\multirow{14}{*}{\centering \rotatebox{90}{mR@100}} 
& MOTIFS~\cite{zellers2018neural}                                  & 15.8 & 13.9 & - & 12.2 & 4.9 & - & 8.5 & 3.4 & - \\ 
& \cc MOTIFS+Ours                       & \cc 16.1 & \cc 14.5 & \cc \textcolor{black}{\bf +4.3\%} & \cc 12.4 & \cc 5.1 & \cc \textcolor{black}{\bf +3.9\%} & \cc 8.4 & \cc 3.5 & \cc \textcolor{black}{\bf +2.9\%} \\
& VCTree~\cite{tang2019learning}          & 16.1 & 14.2 & - & 13.7 & 5.0 & - & 8.6 & 3.1 & - \\ 
& \cc VCTree+Ours                         & \cc 16.5 & \cc 15.0 & \cc \textcolor{black}{\bf +5.6\%} & \cc 14.4 & \cc 5.4 & \cc \textcolor{black}{\bf +8.0\%} & \cc 8.3 & \cc 3.3 & \cc \textcolor{black}{\bf +6.5\%} \\
& VTransE~\cite{tang2020unbiased}         & 15.8 & 13.5 & - & 13.6 & 5.9 & - & 9.2 & 3.6 & - \\ 
& \cc VTransE+Ours                        & \cc 16.1 & \cc 14.4 & \cc \textcolor{black}{\bf +6.7\%} & \cc 13.5 & \cc 6.4 & \cc \textcolor{black}{\bf +8.5\%} & \cc 9.2 & \cc 3.7 & \cc \textcolor{black}{\bf +2.8\%} \\
& HiKER~\cite{zhang2024hiker}             & 41.2 & 35.1 & - & 21.4 & 7.1 & - & - & - & - \\ 
& \cc HiKER+Ours                          & \cc 42.5 & \cc 36.4 & \cc \textcolor{black}{\bf +3.8\%} & \cc 22.3 & \cc 7.6 & \cc \textcolor{black}{\bf +7.1\%} & \cc - & \cc - & \cc - \\
& DPL~\cite{jeon2024semantic}             & 28.6 & 27.6 & - & 20.2 & 7.9 & - & 15.0 & 5.3 & - \\ 
& \cc DPL+Ours                            & \cc 28.7 & \cc 27.9 & \cc \textcolor{black}{\bf +1.1\%} & \cc 21.9 & \cc 8.9 & \cc \textcolor{black}{\bf +12.7\%} & \cc 15.4 & \cc 5.7 & \cc \textcolor{black}{\bf +7.5\%} \\
\bottomrule
\end{tabular}
}
\vspace{-2mm}
\caption{
Comparison of state-of-the-art SGG methods with and without our Robo-SGG module on the clean VG~\cite{krishna2017visual} and corrupted VG-C~\cite{zhang2024hiker} datasets. 
``Corruption Avg.'' denotes the average performance across 20 corruption types at severity level 5. 
Results with Robo-SGG are highlighted in \textcolor{gray}{gray}. 
``Imp.'' reports the relative improvement (in \textbf{bold}) of ``Corrup. Avg.'' over the corresponding baseline. 
The table is divided into two parts: upper for mR@50 and lower for mR@100.
}
\label{Table_main}
\vspace{-5mm}
\end{table*}
\subsection{Training and Inference}

\noindent  \textbf{Loss Function.} Following~\cite{zellers2018neural}, we employ the standard scene graph generation loss function~$\mathcal{L}_{SGG}$, which is the sum of the cross-entropy losses for both predicted objects and predicates. More details are provided in the supplementary materials.

\noindent \textbf{Integration of Robo-SGG into SGG Models.} As a plug-and-play method, the NRM component of Robo-SGG can be applied to the $f_{map}$ obtained from Eq.~(\ref{Eq.1}), while the Object/Predicate LEE primarily affects the Object/Predicate Encoders in Eq.~(\ref{Eq.2}) and Eq.~(\ref{Eq.3}). During training, validation, and testing, Robo-SGG can be seamlessly integrated into the corresponding components of existing SGG models.
\section{Expriments}

\subsection{Experimental Settings}

\textbf{Datasets.}~We use Visual Genome~\cite{krishna2017visual} and GQA~\cite{hudson2019gqa} for both training and testing. To assess robustness, we follow~\cite{zhang2024hiker} and apply 20 types of corruptions to create VG-C and GQA-C, grouped into five classes~\cite{hendrycks2019benchmarking}: Noise (Gaussian, Shot, Impulse), Blur (Defocus, Glass, Motion, Zoom), Weather1 (Snow, Frost, Fog), Digital (Brightness, Contrast, Elastic, Pixelate, JPEG), and Weather2 (Sunlight Glare, Water Drop, Wildfire Smoke, Rain, Dust). More details and visualizations are in the supplementary materials.

\noindent {\bf Metrics.} For evaluation, we evaluate three popular SGG tasks: 1) Predicate Classification (PredCls), 2) Scene Graph Classification (SGCls), and 3) Scene Graph Detection (SGDet). To address the pronounced long-tail distribution in the datasets, we adopt the class-balanced mean recall (mR@K) for the top-K predictions per image. We report results for K = 20, 50, and 100.

\noindent {\bf Compared Methods.} To demonstrate the effectiveness of our model-agnostic approach, we integrate Robo-SGG into two categories of SGG models: (1) three baseline two-stage models: MOTIFS~\cite{zellers2018neural}, VCTree~\cite{tang2019learning}, and VTransE~\cite{tang2020unbiased}; (2) recent state-of-the-art methods: DPL~\cite{jeon2024semantic} and HiKER-SGG~\cite{zhang2024hiker}, the latter being specifically designed for robust SGG. Additionally, we include the one-stage model RelTR~\cite{cong2023reltr} and EGTR~\cite{im2024egtr} for comprehensive evaluation. Notably, following the benchmark protocol in~\cite{zhang2024hiker}, all models~(including baselines and ours) are trained on clean images and evaluated on unseen corrupted images.

\noindent \textbf{Implementation Details.} All experiments are conducted using PyTorch~\cite{paszke2019pytorch} on V100 GPUs. For MOTIFS~\cite{zellers2018neural}, VCTree~\cite{tang2019learning}, and VTransE~\cite{tang2020unbiased}, we follow the Scene-Graph-Benchmark~\cite{tang2020unbiased} settings. For DPL~\cite{jeon2024semantic}, HiKER~\cite{zhang2024hiker}, RelTR~\cite{cong2023reltr}, and EGTR~\cite{im2024egtr}, we use the official code. Further details are provided in the supplementary materials.

\begin{figure*}[h]
\begin{minipage}[t]{1\linewidth}
    \flushright
    \includegraphics[width=0.99
    \linewidth]{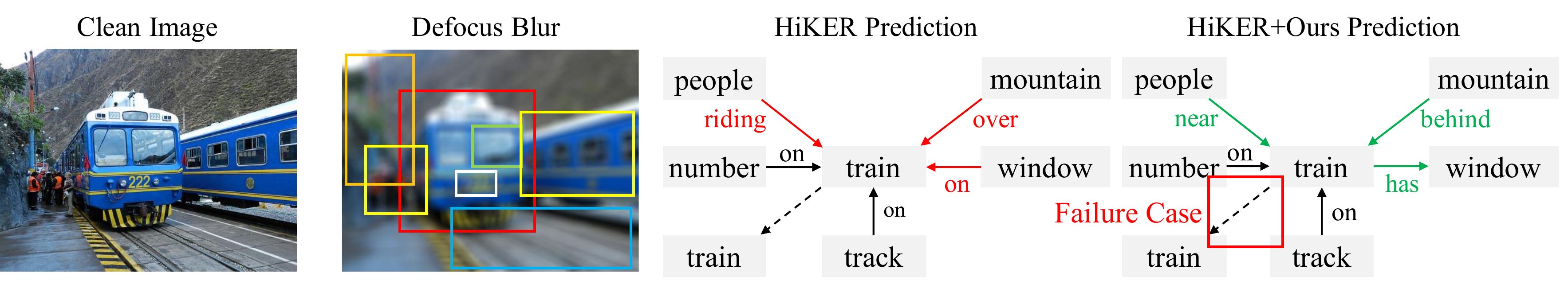}
    \end{minipage}
    \vspace{-8mm}
    \caption{Qualitative comparisons on the PredCls task. Dashed lines: undetected predicates; solid black lines: correct predictions. Red edges: HiKER-SGG errors~\cite{zhang2024hiker}; green edges: our correct predictions.}
    \label{fig: Qualitative comparisons}
    \vspace{-5mm}
\end{figure*} 
\subsection{Results}
\begin{table*}[]
\centering
\resizebox{\textwidth}{!}{
\begin{tabular}{l|cc|cc|cc|cc}
\toprule
Task  & \multicolumn{4}{c}{PredCls}                                             & \multicolumn{4}{c}{SGCls}                                   \\
\midrule
Metric   & \multicolumn{2}{c}{mR@50} & \multicolumn{2}{c}{mR@100}                  & \multicolumn{2}{c}{mR@50}    & \multicolumn{2}{c}{mR@100}   \\
\midrule
Method   & Clean& Corrup. Avg. (Imp.)  & Clean & Corrup. Avg. (Imp.)    & Clean & Corrup. Avg. (Imp.)  & Clean& Corrup. Avg. (Imp.)  \\
\midrule
MOTIFS   & 14.6   & 12.6/ -          & 15.8                 & 13.9/ -              & 11.3   & 4.5/-  &  12.2   &    4.9/-                  \\
$+\text{LEE}$     & 14.5   & \textcolor{black}{\bf 13.0/ ( +3.7\%)}    & 15.6                 & \textcolor{black}{\bf14.1/ ( +1.4\%)}        &  11.3     & \textcolor{black}{\bf4.6/ ( +2.2\%)} &   12.3    &  \textcolor{black}{\bf5.0/ ( +2.0\%)}                    \\
$+\text{NRM}$ & 14.4  & \textcolor{black}{\bf12.8/ ( +2.0\%)}    &  15.5                & \textcolor{black}{\bf14.1/ ( +1.3\%)}        &   11.3    & \textcolor{black}{\bf4.7/ ( +3.7\%)}     &   12.2    & \textcolor{black}{\bf5.0/ ( +1.9\%)}                     \\
$+\text{LEE}+\text{SNR}$ & 15.0   & \textcolor{black}{\bf12.9/ ( +2.4\%)}    & 16.2                 & \textcolor{black}{\bf14.2/ ( +2.2\%)}        & 11.4      & \textcolor{black}{\bf4.6/ ( +2.4\%)}     & 12.4      & \textcolor{black}{\bf5.0/ ( +2.1\%)}                     \\\rowcolor{gray!20}
$+\text{LEE}+\text{NRM}$ & 15.3   & \textcolor{black}{\bf13.8/ ( +9.5\%)}    & 16.0                 & \textcolor{black}{\bf14.5/ ( +4.3\%)}        & 11.4 &    \textcolor{black}{\bf4.9/ ( +8.8\%)}  &    12.4    &     \textcolor{black}{\bf5.1/ ( +3.9\%)}               \\
\midrule
HiKER~\cite{zhang2024hiker}    & 39.3   & 32.6/-           & 41.2                 & 35.1/-               &  20.3     &   3.5/-                   &   21.4    &  7.1/-                     \\
$+\text{LEE}$     & 39.5       &   \textcolor{black}{\bf 32.7/ ( +0.3\%)}   &   41.5   & \textcolor{black}{\bf35.3/ ( +0.5\%)} &   20.5    &   3.5/ ( +0.0\%)                  &  21.5     &7.1/ ( +0.0\%) \\
$+\text{NRM}$     & 39.5       &   \textcolor{black}{\bf 33.5/ ( +2.8\%)}   &   41.8   & \textcolor{black}{\bf35.9/ ( +2.3\%)} &   20.4    &   \textcolor{black}{\bf3.5/ ( +0.6\%)}                 &  21.5     & \textcolor{black}{\bf7.3/ ( +2.7\%)} \\
$+\text{LEE}+\text{SNR}$ & 39.4       &  \textcolor{black}{\bf33.2/ ( +1.8\%)}   &    41.6    & \textcolor{black}{\bf35.8/ ( +2.0\%)} &   20.9    &  \textcolor{black}{\bf3.6/ ( +2.9\%)}                    &   21.8   & \textcolor{black}{\bf 7.3/ ( +2.8\%)} \\\rowcolor{gray!20}
$+\text{LEE}+\text{NRM}$ & 40.8   & \textcolor{black}{\bf33.8/ ( +3.7\%)}    & 42.5   & \textcolor{black}{\bf36.4/ ( +3.8\%)}  &    21.5  &    \textcolor{black}{\bf3.7/ ( +5.7\%)}                & 22.3       &  \textcolor{black}{\bf7.6/ ( +7.1\%)} \\
\bottomrule
\end{tabular}
}
\vspace{-2mm}
\caption{{Performance comparison of the state-of-the-art SGG methods with integration of our method for the PredCls and SGCls task.} ``Corrup. Avg.'' and ``Imp'' denote the average performance under different corruption conditions and the performance improvement of our Robo-SGG over the baseline under corruption, respectively, highlighted in {\bf bold}.}

\label{table8}
\vspace{-6mm}
\end{table*}

\begin{table}[h]
\small
\begin{tabular}{l|ccc}
\toprule
\multirow{2}{*}{Method} & \multicolumn{3}{c}{GQA-200(mR@20)} \\
                        & Clean   & Corruption Avg.  & Improvement(\%)      \\
\midrule
MOTIFS               & 20.5    & 16.0           & -       \\
$+\text{LEE} $             & 21.6    & 16.3         & \textcolor{black}{\bf +1.9\%}   \\
$+\text{LEE}+\text{SNR}$           & 21.5    & 17.5         & \textcolor{black}{\bf+9.4\%}   \\ \rowcolor{gray!20}
$+\text{LEE}+\text{NRM} $              & 21.7    & 17.8         & \textcolor{black}{\bf+11.3\%}  \\
\midrule
VCTree                & 21.0      & 16.5         & -      \\
$+\text{LEE} $             & 21.6    & 17.3         & \textcolor{black}{\bf+4.8\%}   \\
$+\text{LEE}+\text{SNR} $            & 21.5    & 17.5  & \textcolor{black}{\bf+6.1\%}   \\ \rowcolor{gray!20}
$+\text{LEE}+\text{NRM} $             & 21.8    & 17.9         & \textcolor{black}{\bf+8.5\%}  \\
\midrule
VTransE            & 22.5    & 17.4         & -       \\
$+\text{LEE} $              & {23.0}    & 18.4         & \textcolor{black}{\bf+5.7\%}   \\
$+\text{LEE}+\text{SNR} $              & 22.9    & 18.6         & \textcolor{black}{\bf+6.9\%}   \\ \rowcolor{gray!20}
$+\text{LEE}+\text{NRM} $            & 22.9    & {19.1}         & \textcolor{black}{\bf +9.8\%}  \\
\bottomrule
\end{tabular}
\vspace{-2mm}
\caption{{ Performance comparison of the SGG methods with integration of our method for the PredCls task on GQA~\cite{hudson2019gqa} dataset.}}
\label{table9}
\vspace{-3mm}
\end{table}

\begin{figure}
    \centering
    \includegraphics[width=1.0\linewidth]{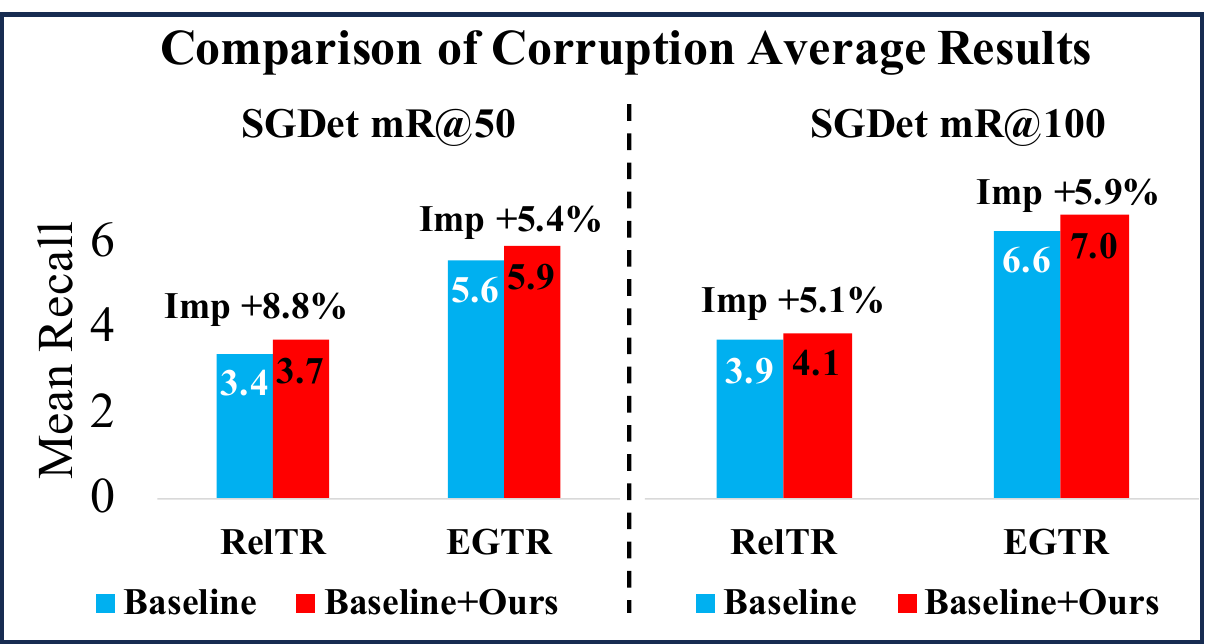}
    \vspace{-7mm}
    \caption{Comparison of state-of-the-art SGG methods with and without our Robo-SGG.}
    \label{fig: one stage}
    \vspace{-7mm}
\end{figure}

\begin{table}[t]
\small
\begin{tabular}{l|ccc}
\toprule
Method & Clean & Corruption Avg. & Improvement (\%) \\
\midrule
VCTree                         & 7.32 & 2.51 & -- \\
$+\text{LEE}_{\text{Add}}$      & 7.33 & 2.48 & -1.2 \\
$+\text{LEE}_{\text{Concat}}$   & 7.33 & 2.46 & -2.0 \\
\rowcolor{gray!20}
$+\text{LEE}_{\text{Gate}}$     & 7.36 & 2.62 & \textbf{+4.2} \\
\midrule
$+\text{NRM}_{\text{bbox}}$     & 7.29 & 2.55 & +1.6 \\
\rowcolor{gray!20}
$+\text{NRM}_{\text{centroid}}$ & 7.33 & 2.69 & \textbf{+7.2} \\

\bottomrule
\end{tabular}
\vspace{-3mm}
\caption{Ablation study on different fusion strategies of LEE and different layout-aware attention of NRM (SGDet mR@50).}
\vspace{-5mm}
\label{tab:fusion_ablation}
\end{table}

\begin{table}[t]
\small
\begin{tabular}{l|ccc}
\toprule
Setting & Clean & Corruption Avg. & Improvement (\%) \\
\midrule
\rowcolor{gray!20}
\multicolumn{4}{c}{\textbf{\textit{30\% Bbox Perturbed}}} \\
\quad MOTIFS & 6.231 & 2.810 & -- \\
\quad +Ours  & 6.235 & 2.865 & \textbf{+2.0} \\
\midrule
\rowcolor{gray!20}
\multicolumn{4}{c}{\textbf{\textit{Style Change}}} \\
\quad MOTIFS & 7.051 & 2.381 & -- \\
\quad +Ours  & 7.064 & 2.682 & \textbf{+12.6} \\
\midrule
\rowcolor{gray!20}
\multicolumn{4}{c}{\textbf{\textit{Distribution Shift (Zero-shot)}}} \\
\quad MOTIFS & 0.509 & 0.245 & -- \\
\quad +Ours  & 0.515 & 0.288 & \textbf{+17.8} \\
\bottomrule
\end{tabular}
\vspace{-3mm}
\caption{Robustness under different settings (SGDet mR@50).}
\vspace{-6mm}
\label{tab:robustness}
\end{table}

\noindent \textbf{Quantitative Results.}
Table~\ref{Table_main} reports results for the three SGG tasks, with models trained on VG and tested on the unseen corruptions in VG-C. After integrating Robo-SGG, all SGG models show less performance degradation in mR@50 and mR@100 across PredCls, SGCls, and SGDet tasks, which demonstrates the robustness of our model in handling corrupted scenarios. For example, with the classic MOTIFS~\cite{zellers2018neural}, the integration of Robo-SGG reduces the performance drops by 0.6, 0.4, and 0.1 in mR@50 for PredCls, SGCls, and SGDet, respectively. Similar improvements also appear for VCTree~\cite{tang2019learning} and VTransE~\cite{tang2020unbiased}. Compared to HiKER~\cite{zhang2024hiker}, which is specifically designed for robust SGG under corruptions, our method achieves improvements of 3.7\% and 5.7\% in mR@50 for PredCls and SGCls, respectively. Furthermore, on the latest state-of-the-art model DPL~\cite{jeon2024semantic}, our method achieves relative gains of 6.3\% and 7.5\% in mR@50 and mR@100 for the more challenging SGDet task. We also evaluate Robo-SGG on the one-stage model in Figure~\ref{fig: one stage}. RelTR~\cite{cong2023reltr}, where the SGDet mR@50 under corruption improves from 3.4 to 3.7, corresponding to an 8.8\% relative gain. While EGTR's performance in SGDet mR@50 is 17.8/5.6 (clean/corruption), the performance with EGTR+ours improved to 17.8/5.9, achieving a 5.4\% relative improvement under VG-C datasets. We further validate the stability of our gains by running MOTIFS+Ours (SGCls) with three random seeds, yie lding a standard deviation of less than 0.01, much smaller than the reported improvement (e.g., +11.1\% in SGCls).

\noindent \textbf{Qualitative Results.}~To further demonstrate our method's effectiveness, Figure~\ref{fig: Qualitative comparisons} shows scene graphs generated by HiKER~\cite{zhang2024hiker} and HiKER with Robo-SGG on corrupted images. For example, defocus blur visually obscures the boundaries between different objects and reduces the distinction between predicates, causing HiKER to incorrectly predict the triplet as $\langle$people-riding-train$\rangle$. In contrast, our method leverages NRM to alleviate the impact of blur on feature maps and utilizes the layout information to recover the structural features between ``people'' and ``train'', resulting in the correct prediction $\langle$people-near-train$\rangle$. We also observe a failure case, possibly because both ``near'' and ``behind'' can describe the relationship, which could be further addressed with more fine-grained annotations. More examples appear in the supplementary materials.

\subsection{Ablation Study}
\begin{figure}[h]
    \centering
    \includegraphics[width=0.8\linewidth]{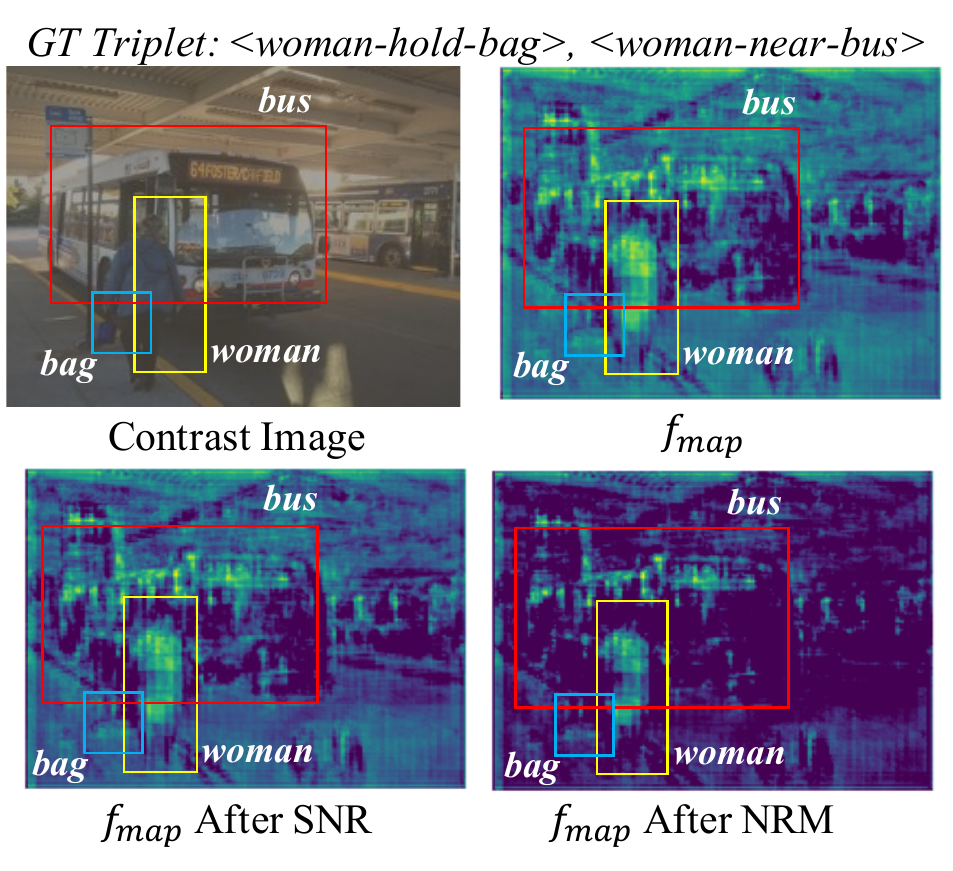}
    \vspace{-4mm}
   \caption{{ Visualization of corrupted image and its feature maps.} The regions of ``woman'', ``bag'', and ``bus'' are highlighted with boxes. ``GT'' denotes Ground Truth.}
    \label{fig:feature map}
     \vspace{-7mm}
\end{figure}

{\bf Layout-Oriented Normalization and Restitution Module~(NRM).} Table~\ref{table8} compares our NRM with the SNR method~\cite{jin2020style}, which uses channel attention for feature restitution. While SNR improves robustness under corruption (\textit{e.g.}, +2.2\% and +2.0\% in PredCls mR@100 on MOTIFS~\cite{zellers2018neural} and HiKER~\cite{zhang2024hiker}), our NRM achieves even greater and more consistent gains (+4.3\% and +3.8\%), demonstrating superior effectiveness in recovering structural features for relationship recognition. Figure~\ref{fig:feature map} further visualizes feature maps under corruption. For example, when contrast corruption causes SNR to focus on irrelevant background, NRM suppresses background noise and enhances structural features between objects (\textit{e.g.}, woman-bus, woman-bag), enabling correct predictions such as $\langle$woman-hold-bag$\rangle$ and $\langle$woman-near-bus$\rangle$. This demonstrates that our Layout-aware Attention more effectively restores structural features essential for scene graph generation. We also ablate the spatial input to the layout-aware attention: using object centroids (\(\text{NRM}_{\text{centroid}}\)) outperforms using full bounding boxes (\(\text{NRM}_{\text{bbox}}\)), as centroids are less sensitive to detection noise.

\noindent {\bf Layout-Embedded Encoder~(LEE).}  
Table~\ref{table8} shows the impact of Object/Predicate LEE on VG~\cite{krishna2017visual} and VG-C~\cite{zhang2024hiker}. Key findings: 1) On clean VG, adding LEE brings a 0.2\% mR@50 gain in both PredCls and SGCls for HiKER~\cite{zhang2024hiker}. 2) On VG-C, LEE yields larger improvements, with MOTIFS~\cite{zellers2018neural} increasing by 3.7\% in PredCls and 2.2\% in SGCls. Similar results are observed on GQA and GQA-C (Table~\ref{table9}), where both baselines benefit from LEE and NRM. 3) As Table~\ref{tab:fusion_ablation} shows, directly concatenating or adding bounding box embeddings ($\text{LEE}_{\text{Concat}}$ and $\text{LEE}_{\text{Add}}$) under corruption degrades performance due to unreliable detections. In contrast, our gate fusion adaptively downweights noisy box cues and balances visual and spatial information, thus improving robustness in relationship detection and classification. Further analysis reveals that the gating coefficient $z_i$ in Eq.~\eqref{Eq.17} dynamically shifts with corruption severity. Under Gaussian noise, the mean $\mathbb{E}[z_i]$ decreases from 0.65 (severity 1) to 0.52 (severity 5), indicating that LEE increasingly relies on layout cues as visual quality deteriorates. Moreover, when object bounding boxes are perturbed by random noise (±30\% of width and height), our method still improves the baseline by +2.0\% (Table~\ref{tab:robustness}), demonstrating the effectiveness of the gating mechanism in leveraging global layout structure even under inaccurate detection.

\section{How does Robo-SGG help?}

\noindent {\bf Runtime and Memory Overhead.} Table~\ref{tab:memory_comparison} shows that LEE and NRM increase inference time by only 0.005s and 0.019s, respectively, on SGCls. Notably, NRM introduces no extra learnable parameters and does not increase GPU memory usage. LEE adds only 0.02GB of memory during inference. Both modules together achieve an 8.8\% mR@50 improvement, which demonstrates a favorable trade-off.

\begin{table}[h]
\small
\centering
\vspace{-2mm}
\begin{tabular}{lcccc}
\toprule
\multirow{2}{*}{Method} & \multicolumn{2}{c}{Training} & \multicolumn{2}{c}{Inference} \\
\cmidrule(lr){2-3} \cmidrule(lr){4-5}
 & Time/iter & Memory & Time/iter & Memory \\
\midrule
MOTIFS        & 0.777s & 13.57GB & 0.100s & 7.28GB \\
+LEE          & 0.800s & 16.56GB & 0.105s & 7.30GB \\
+NRM          & 1.036s & 13.57GB & 0.119s & 7.28GB \\
+LEE+NRM      & 1.070s & 13.56GB & 0.127s & 7.30GB \\
\bottomrule
\end{tabular}
\vspace{-3mm}
\caption{Cost analysis (per iteration) for SGCls task.}
\vspace{-3mm}
\label{tab:memory_comparison}
\end{table}

\begin{figure}[t]
    \centering
    \includegraphics[width=0.9\linewidth]{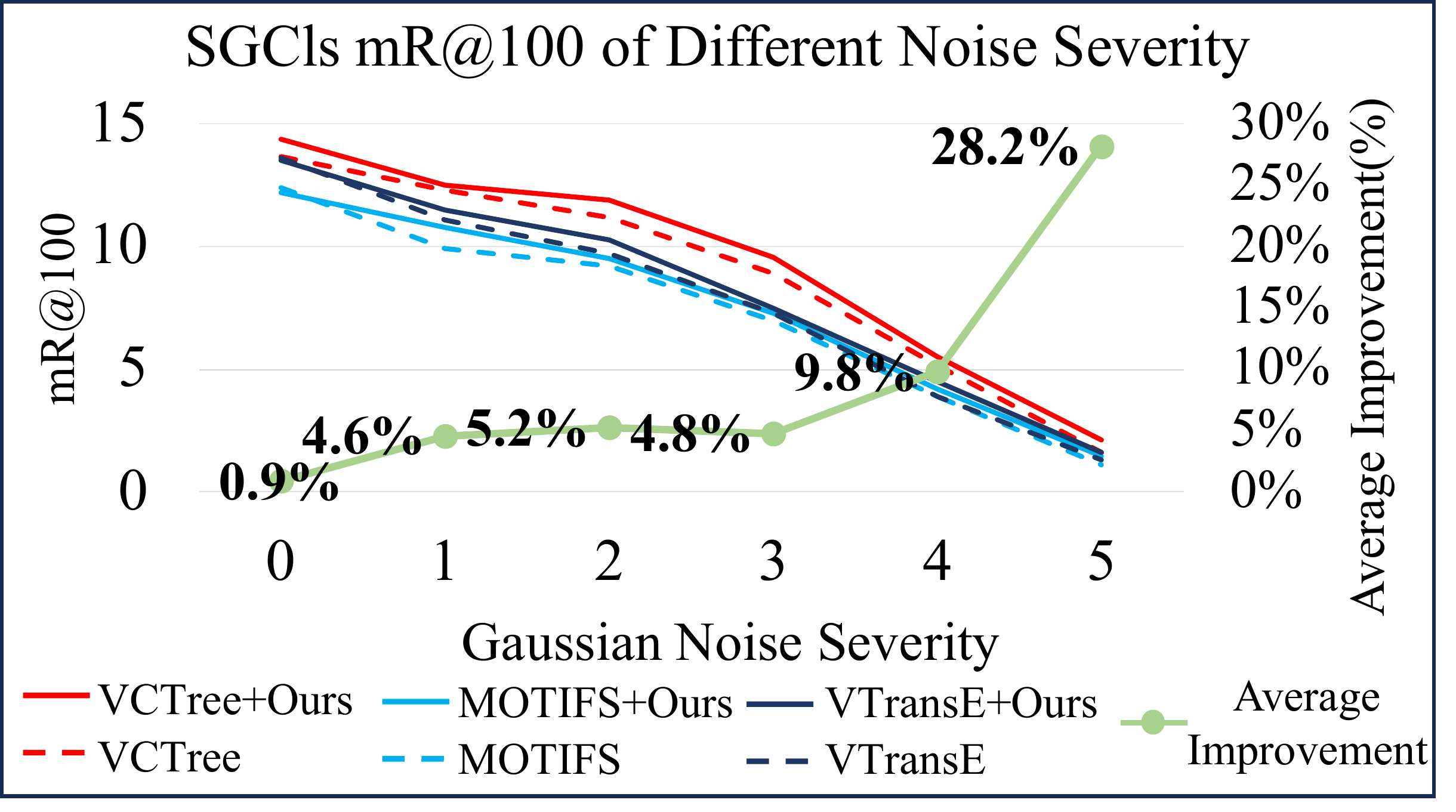}
    \vspace{-3mm}
    \caption{Comparison of baseline and our Robo-SGG under different levels of Gaussian noise.}
    \label{fig: fold}
    \vspace{-5mm}
\end{figure}
\noindent{\bf Analysis of Corruption Severity.}
We evaluate mR@100 in the SGCls task under five increasing levels of corruption severity~\cite{hendrycks2019benchmarking} (see Figure~\ref{fig: fold}). As the severity level increases, all methods exhibit performance degradation; however, Robo-SGG consistently outperforms the baseline across all levels. Notably, the relative improvement of our method progressively grows, from 4.6\% at severity level 1 to 28.2\% at level 5, demonstrating its increasing effectiveness under more challenging conditions. This trend primarily stems from the NRM’s ability to leverage domain-invariant global layout structures to effectively recover robust structural features even under severe corruption.

\noindent{\bf Analysis of Other Domain Shift.}
As shown in Table~\ref{tab:robustness}, we further evaluate robustness under other domain shifts. Specifically, we use StyleID~\cite{chung2024style} to alter image styles, representing another type of domain shift, and also test zero-shot scenarios with mismatched label distributions~(Distribution Shift). In both cases, our model achieves strong improvements over MOTIFS, demonstrating its robustness to various distribution changes.

\begin{table}[t]
\centering
\small
\setlength{\tabcolsep}{4pt} 
\begin{tabular}{lcc|cc}
\toprule
Method & SGCls$\uparrow$ & Imp.(\%) & Inference Time$\downarrow$ & Imp.(\%) \\
\midrule
HiKER & 1.5 & -- & 1.75s/image & -- \\
+DDPG~\cite{garber2024image} & 1.7 & 13.3 & 43.7s/image & 2394.3 \\
+Ours & 2.2 & 46.7 & 1.79s/image & 2.3 \\
\bottomrule
\end{tabular}
\vspace{-3mm}
\caption{Comparison between DDPG and our method on motion-blurred images (SGCls mR@100).}
\vspace{-5mm}
\label{table10}
\end{table}

\noindent{\bf Comparison with Denoising Method.}
Existing methods such as DDPG~\cite{garber2024image} address image corruptions like motion blur. We use its pre-trained model to denoise images before SGG, running all experiments on a single Tesla 32 GB V100 GPU. As shown in Table~\ref{table10}, DDPG-denoised images improve performance by 13.3\%, but inference time increases by 2394.3\%. In contrast, our method achieves a 46.7\% gain with only a 2.3\% increase in inference time. This is because diffusion-based denoising, while visually appealing, often removes high-frequency details and produces overly smooth images~\cite{whang2022deblurring}, whereas our NRM preserves structural features via Layout-aware Attention for better robustness.

\noindent {\bf Per-corruption Detailed Results.}
Our method improves performance across all 20 corruption types, with the largest gains under Noise (Gaussian +31.1\%, Shot +28.8\%, Impulse +58.7\%) and Digital (Pixelate +13.3\%, JPEG +63.7\%) corruptions. Blur-related degradations (e.g., Defocus +9.6\%, Zoom +4.8\%) show more limited gains, as blur obscures object boundaries and degrades structural cues. Full results are in the supplementary material.

\noindent {\bf Error Analysis under Corruption.}
In SGDet under motion blur, baselines relying solely on distorted visual features may mispredict $\langle$man–wearing–shirt$\rangle$ as $\langle$man–on–shirt$\rangle$, while those overly dependent on bounding boxes tend to output $\langle$man–in–shirt$\rangle$. Our Robo-SGG adaptively fuses both cues, correcting these errors and recovering the correct predicate “wearing”. Full analysis is in the supplementary material.
\section{Conclusion}
In this paper, we attribute corruption’s impact on scene graph generation to domain shift in image features. We propose Robo-SGG, a plug-and-play module for robust scene graph generation, consisting of the Layout-Oriented Normalization and Restitution Module for image features and the Layout-Embedded Encoder for object and predicate encoding. Our method easily integrates into existing models and consistently improves multiple baselines. Extensive experiments on VG-C and GQA-C benchmarks demonstrate its effectiveness and state-of-the-art performance. 
{
    \small
    \bibliographystyle{ieeenat_fullname}
    \bibliography{main}
}


\end{document}